\documentclass[10pt,twocolumn,letterpaper]{article}

\usepackage{wacv}
\usepackage{times}
\usepackage{epsfig}
\usepackage{graphicx}
\usepackage{amsmath}
\usepackage{amssymb}
\usepackage{subcaption}

\def\wacvPaperID{****} 

\ifwacvfinal
\def\assignedStartPage{9876} 
\fi

\ifwacvfinal
\usepackage[breaklinks=true,bookmarks=false]{hyperref}
\else
\usepackage[pagebackref=true,breaklinks=true,colorlinks,bookmarks=false]{hyperref}
\fi

\ifwacvfinal
\else
\fi

\begin{document}

\title{A Comparative Analysis of the Face Recognition Methods in Video Surveillance Scenarios}

\author{Onur Eker\\
Havelsan Inc., Ankara, Turkey\\
{\tt\small onureker@hacettepe.edu.tr}
\and
Murat Bal\\
Havelsan Inc., Ankara, Turkey\\
{\tt\small muratbal@hacettepe.edu.tr}
}

\maketitle

\begin{abstract}
Facial recognition is fundamental for a wide variety of security systems operating in real-time applications. In video surveillance based face recognition, face images are typically captured over multiple frames in uncontrolled conditions; where head pose, illumination, shadowing, motion blur and focus change over the sequence. We can generalize that the three fundamental operations involved in the facial recognition tasks: face detection, face alignment and face recognition. This study presents comparative benchmark tables for the state-of-art face recognition methods by testing them with same backbone architecture in order to focus only on the face recognition solution instead of network architecture. For this purpose, we constructed a video surveillance dataset of face IDs that has high age variance, intra-class variance (face make-up, beard, etc.) with native surveillance facial imagery data for evaluation. On the other hand, this work discovers the best recognition methods for different conditions like non-masked faces, masked faces, and faces with glasses.

\end{abstract}

\section{Introduction}

Face recognition is a well-established research problem in the field of computer vision with the aim of recognizing human identities by facial images. Face recognition is recognized as one of the most important tools for a wide variety of identity applications, from law enforcement and information security to business, entertainment and e-commerce. One of the driving forces behind the success of face recognition is the rapid advances in deep learning techniques, large face recognition benchmarks, and powerful computing devices. On the major benchmarks (such as LFW~\cite{huang2008labeled}, MegaFace~\cite{kemelmacher2016megaface}), face recognition accuracy on good quality images has reached an unprecedented level thanks to deep learning. However, success of these methods do not scale to native low resolution face image data captured from surveillance videos in unrestricted environments.

\textbf{Challenges of Face Recognition in Video Surveillance}. Low quality and resolution of video frames, makes the recognition accuracy to decrease drastically. Often individuals passes by walking from the camera view, appearing on many frames. They do not look to camera directly which decrease the number of frontal face samples and many of these face samples may not be suitable for face recognition. In surveillance videos, limited control of acquisition conditions such as variation in poses, expressions, illumination, cooperation of individuals, occlusion or motion blur are some of the factors that make face recognition harder in surveillance videos. We list some of the other challenges of face recognition in surveillance videos below: 

\begin{itemize}
\item Inter-class and intra-class variability and noise in the feature space

\item Ageing and variation of interaction between sensor and individual

\item Facial models are often poor representatives of real faces

\item Highly skewed data distributions: very few positives (from individuals of interest) w.r.t. negative samples (from open world)
\end{itemize}

In this study, we constructed a dataset of identities with varying ages, intra-class variance (face make-up, beard, etc.) with native surveillance facial imagery data for evaluation. Morever, we constructed two synthetic versions of this dataset. In one version we added glasses, in the other version we added face masks to all of the faces appearing in the videos in order to analyze the performance of the methods when face is occluded. We benchmark nine different representative deep learning face recognition models for face identification on these three scenarios with discussions and analysis.

\textbf{Strategies for evaluating face recognition methods}. In order to make a fair comparison between face recognition methods, all of the methods should be trained with the same dataset. On the other hand, it is also crucial to use the same backbone for all face recognition networks. Here we used FaceX-Zoo toolbox~\cite{wang2021facex}, which provides a training module with various state-of-the-art face recognition methods and backbones. Similar to infering standalone images, we just applied face detection and face recognition method for each frame independently without any consideration of face quality, detection or identification information from previous frames.



\begin{table*}[h]
\centering
\caption{The performance (\%) of different methods on popular face recognition benchmarks. where RFW (Afr), RFW (Asi), RFW (Cau) and RFW (Ind) denote the African, Asian, Caucasian and Indian test protocols in RFW. Apart from MegaFace, they report the mean accuracies on these benchmarks. For MegaFace, they report the Rank-1 accuracy. Table is taken from~\cite{wang2021facex}.}
\label{tab:facex_scores}
\resizebox{\textwidth}{!}{
\begin{tabular}{l|lllllllll}
\textbf{Method} & \textbf{LFW} & \textbf{CPLFW} & \textbf{CALFW} & \textbf{AgeDB} & \textbf{RFW(Afr)} & \textbf{RFW(Asi)} & \textbf{RFW(Cau)} & \textbf{RFW(Ind)} & \textbf{MegaFace} \\ \hline
AM-Softmax & 99.58 & 83.63 & 93.93 & 95.85 & 88.38 & 87.88 & 95.55 & 91.18 & 88.92 \\
AdaM-Softmax & 99.58 & 83.85 & 93.50 & 96.02 & 87.90 & 88.37 & 95.32 & 91.13 & 89.40 \\
AdaCos & 99.65 & 83.27 & 92.63 & 95.38 & 85.88 & 85.50 & 94.35 & 88.27 & 82.95 \\
ArcFace & 99.57 & 83.68 & 93.98 & 96.23 & 88.22 & 88.00 & 95.13 & 90.70 & 88.39 \\
MV-Softmax & 99.57 & 83.33 & 93.82 & 95.97 & 88.73 & 88.02 & 95.70 & 90.85 & 90.39 \\
CurricularFace & 99.60 & 83.03 & 93.75 & 95.82 & 88.20 & 87.33 & 95.27 & 90.57 & 87.27 \\
CircleLoss & 99.57 & 83.42 & 94.00 & 95.73 & 89.25 & 88.27 & 95.32 & 91.48 & 88.75 \\
NPCFace & 99.55 & 83.80 & 94.13 & 95.87 & 88.08 & 88.20 & 95.47 & 91.03 & 89.13
\end{tabular}}
\end{table*}

\section{Related Work}
In the literature, the problem of video surveillance face recognition is significantly under-studied when compared to face recognition in web photos. While web photo facial recognition is popular and commercially attractive due to the advancements in social media and e-commerce on smartphones and various digital devices, solving the surveillance face recognition problem is critical for public safety and law enforcement applications. The main reason for not developing robust and scalable facial recognition models suitable for surveillance videos is the lack of large-scale benchmarks of video surveillance face images, in contrast to the rich availability of high-resolution web photography face recognition benchmarks. For example, face recognition performance reached 99.65\% on LFW and 90.39\% on MegaFace as can be seen in Table~\ref{tab:facex_scores}. The table is taken from FaceX-Zoo, which also presents the methods we have evaluated in this study for the task of video surveillance face recognition.  

Because of both restricted data access and very laborious data labeling, generating large-scale video surveillance facial image data as a benchmark for larger research is expensive. Currently, one of the largest surveillance face recognition challenge benchmark is the UnConstrained College Students (UCCS) dataset~\cite{gunther2017unconstrained}, which contains 100k face images from 1,732 face IDs at a significantly smaller scale than the MegaFace dataset. Unfortunately, the dataset is removed and not accessible due to concerns of the students since they are photographed without their knowledge. Another benchmark is the QMUL-SurvFace challenge benchmark~\cite{cheng2018surveillance}, that contains 463,507 face images of 15,573 IDs. QMUL-SurvFace is the largest dataset for surveillance face recognition challenge, with native low-resolution surveillance facial images captured by unconstrained wide-field cameras. The dataset is constructed by data-mining 17 public domain person re-identification datasets using a deep learning face detection model. QMUL-SurvFace only provides face crops that is gathered from the surveillance videos, which prohibits studying on hybrid models that is composed of tracking and re-identification that may increase the performance on video surveillance face recognition. On the other hand, NIST proposed the Face in Video Evaluation (FIVE)~\cite{grother2017face} competition for the face recognition of non-cooperating subjects on six different videos. Sixteen major commercial suppliers of face recognition technologies submitted thirty six algorithms to this competition. Unfortunately, the details of the algorithms and the dataset is not shared. Problems encountered with these benchmarks and datasets motivated us to construct our own surveillance video dataset.

\section{Methods}
\begin{figure}[h]
    \centering
    \includegraphics[width=1\linewidth,height=0.05\textheight]{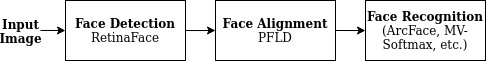}
    \caption{Overview of the evaluation pipeline. }
    \label{fig:pipeline}
\end{figure}

FaceX-Zoo toolbox~\cite{wang2021facex} is used to evaluate and compare the methods on our custom video surveillance dataset. FaceX-Zoo is a PyTorch toolbox for face recognition. It provides a training module with various face recognition methods and backbones. It also provides a model zoo that contains pretrained state-of-the-art face detection and recognition models. In this project, we fixed the face detection and the face alignment methods, and we compared the face recognition performance of different face recognition algorithms in surveillance videos under the same settings. A simple overview of the evaluation pipeline is given in Figure~\ref{fig:pipeline}. 

RetinaFace~\cite{deng2020retinaface} is used as the detection method which is a single-shot, multi-level face detection method. Detected faces are then cropped and aligned with PFLD method~\cite{guo2019pfld} which is an efficient and effective facial landmark detector, that consists of a backbone network and an auxiliary network. Next, the aligned faces are given as inputs to different recognition networks to evaluate the performances of these recognition methods. We fixed all of the recognition algorithms backbone network to fairly compare the methods. Here we used MobileFaceNet \cite{howard2017mobilenets} architecture as the backbone network to extract the features of the aligned faces. MobileFaceNet is a highly efficient network for the application on mobile devices, which is very crucial for video-surveillance.

There are two major research directions to train DCNNs for face recognition. Some methods train a multi-class classifier that can separate the different identities in the training set using a softmax classifier~\cite{wang2018additive}, some other methods try to directly learn an embedding~\cite{schroff2015facenet}. Here, we compare the performance of different state-of-the-art softmax-based face recognition methods. In the softmax-based recognition methods, to learn the discriminative features for face recognition, the estimated logits are generally processed with some operations such as normalization, scaling, and adding margin before the softmax layer. FaceX-Zoo implements a series of softmax style losses in as follows~\cite{wang2021facex}:

\begin{itemize} 
\item AM-Softmax~\cite{wang2018additive}: An additive margin loss that adds a cosine margin penalty to the target logit.

\item ArcFace~\cite{Deng_2019_CVPR}: An additive angular margin loss that adds a margin penalty to the target angle.

\item AdaCos~\cite{zhang2019adacos}: A cosine-based softmax loss that is hyperparameter-free and adaptive scaling.

\item AdaM-Softmax~\cite{liu2019adaptiveface}: An adaptive margin loss that can adjust the margins for different classes adaptively.
 
\item CircleLoss~\cite{sun2020circle}: A unified formula that learns with class-level labels and pair-wise labels.

\item CurricularFace~\cite{huang2020curricularface}: A loss function that adaptively adjusts the importance of easy and hard samples during different training stages.

\item MV-Softmax~\cite{wang2020mis}: A loss function that adaptively emphasizes the misclassified feature vectors to guide the discriminative feature learning.

\item NPCFace~\cite{zeng2020npcface}: A loss function that emphasizes the training on both the negative and positive hard cases.
 
\item MagFace~\cite{meng2021magface}: An adaptive mechanism introduced to ArcFace in order to learn a well-structured within-class feature distributions.
\end{itemize}

\begin{figure*}[h]
\begin{center}
    \includegraphics[width=0.3\linewidth,height=0.15\textheight,trim=0 6.6cm 0 1cm, clip]{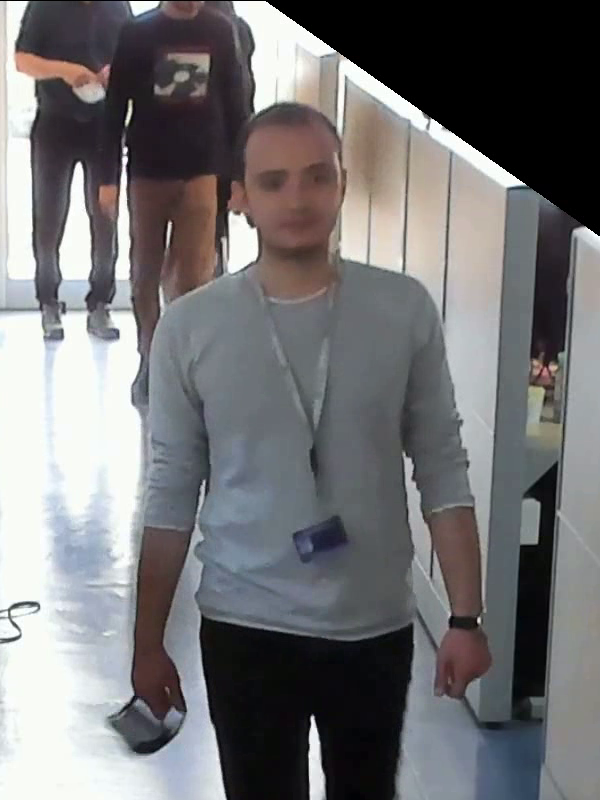}
    \includegraphics[width=0.3\linewidth,height=0.15\textheight,trim=0 6.6cm 0 1cm, clip]{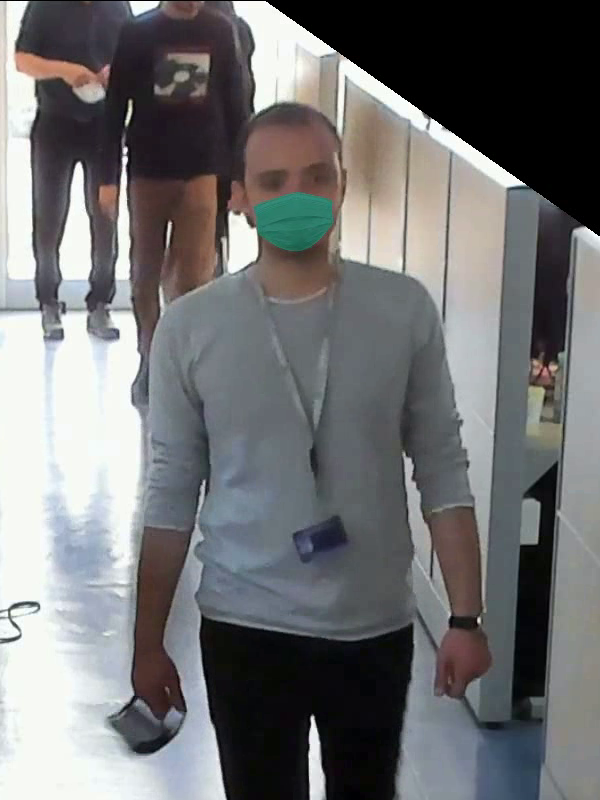}
    \includegraphics[width=0.3\linewidth,height=0.15\textheight,trim=0 6.6cm 0 1cm, clip]{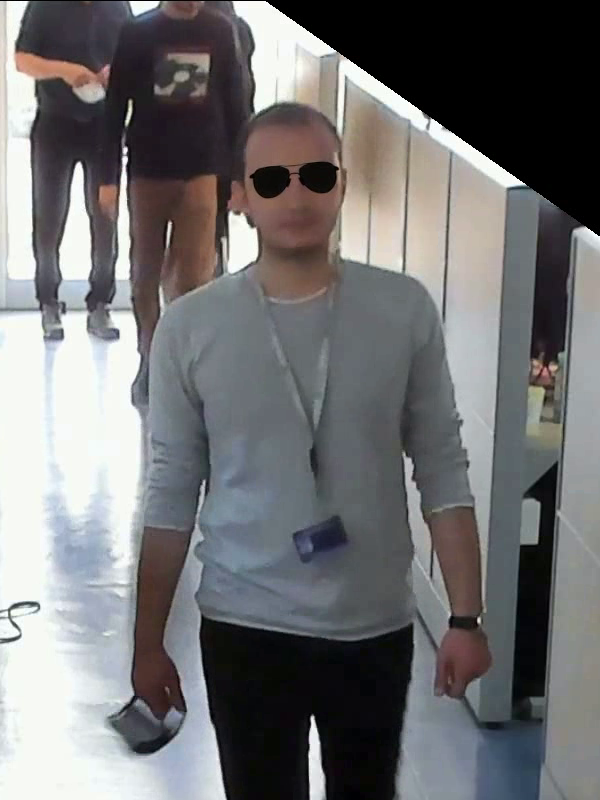}
\end{center}
\caption{Sample frame from a query video: clean face, face with mask, and face with glasses, respectively. }
\label{fig:sample_frames}
\end{figure*}

\section{Video Surveillance Dataset}

Although there are quite a number of datasets in the literature to measure face recognition performance from images~\cite{guo2016ms,kemelmacher2016megaface}, the number of surveillance video datasets that are more suitable for measuring performance in real-world scenarios is quite limited as mentioned previously. Because of this, we created our own surveillance dataset which consist of sequences that contains different illuminations, partial occlusions and head poses of different male and female subjects. In video sequences, the faces of subjects in the camera view are between approximately $20\times20$ pixels and $400\times400$ pixels. Very poor quality regions are considered as don't care regions and excluded from the performance measurements. These don't care regions is decided for image regions that cannot be identified by human perception. The query data is first automatically annotated with CVAT annotation tool~\cite{cvat} and then refined manually.

\begin{table}[]
\centering
\caption{The statistics of the custom video surveillance dataset.}
\label{tab:dataset_stats}
\resizebox{\columnwidth}{!}{
\begin{tabular}{l|c}
\textbf{Property} & \textbf{Statistics} \\ \hline
Min. bounding box & W=12 - H=25 \\
Max. bounding box & W=349 - H=459 \\
Mean bounding box & W=95 - H=123 \\
\# of frames in the query set & 33630 \\ 
\# of face boxes & 7143 \\
\# of known identities in the query set & 161 \\
\# of unknown identities in the query set & 110 \\
\# of people enrolled in the gallery & 2627
\end{tabular}}
\end{table}

\begin{figure}[h]
    \centering
    \includegraphics[width=0.8\linewidth,height=0.18\textheight]{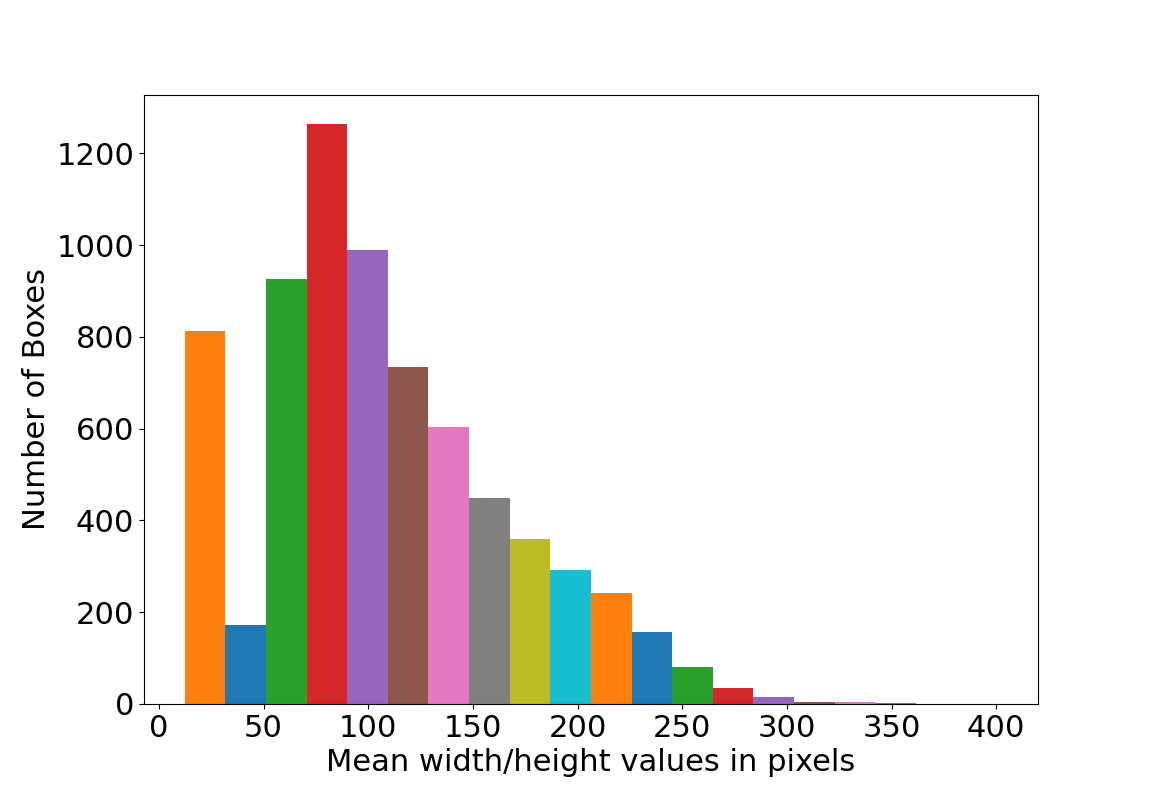}
    \caption{Histogram of mean width/height values of the bounding boxes in pixels. }
    \label{fig:box_hist}
\end{figure}

Synthetic face datasets are generated by adding masks and glasses to the face images synthetically. Figure~\ref{fig:sample_frames} shows a sample frame from query dataset. As can be seen from the figure, we created two additional dataset using the clean dataset, which are the faces with masks and faces with glasses. Image resolution of the video sequences is $1080\times1920$. Table~\ref{tab:dataset_stats} shows statistics of the dataset. Number of identities that enrolled in the gallery (2627) is far more than number of known identities (161) in the query videos. This is often referred to the watch-list identification scenario where only persons of interest are enrolled into the gallery, and the others are distractors. There are also 110 unknown identities in the query videos, which are not enrolled in the gallery. There is a high intra-class variance in the dataset such as varying age-gaps, facial make-ups, beard, etc. For example, gallery image of a person could be captured at the age of 20, while the person is at the of 40 in the query set. Bounding box sizes of the faces varies from $20\times20$ to $400\times400$ with a mean bounding box size of $95\times123$. Figure~\ref{fig:box_hist} shows the distribution of the size of the face bounding boxes. The dataset mostly contains low-resolution face images which is coherent with real-world surveillance data. There are 33630 frames in the dataset and 7143 face bounding boxes of the identities in the query set. Unfortunately, due to the privacy and security policies employed by our company, we are unable to provide more representative samples and details of the dataset.

\section{Experiments}
We use the pretrained models provided by FaceX-Zoo toolbox to compare and evaluate the methods on our custom video surveillance dataset. All of the methods are trained with the same setting, which is crucial for a fair comparison. They use MS-Celeb-1M-v1c~\cite{Deng_2019_CVPR} as the training data and adopt the same backbone, MobileFaceNet. Four NVIDIA Tesla P40 GPUs are employed for training. They trained the methods for 18 epochs and set the batch size to 512. The learning rate is initialized as 0.1, and divided by 10 at the epoch 10, 13 and 16.

Matching performances of the methods in the video sequences is measured separately for each scenario and then average score is calculated. Face recognition performance of the methods for the video sets is calculated in two different ways: the matching performance of correctly matching faces in the video set with people enrolled in the gallery, the separation performance of the methods when faces in the video set are not enrolled in the gallery. In case the identified faces are enrolled in the gallery, similar to Cumulative Matching Characteristic (CMC) metric~\cite{jain2015guidelines}, for a face query list, the face matching performance is calculated for each matching level $k$ in a predefined rank $R$. In case the identified faces are not enrolled in the gallery the separation performance, which measures detecting an unknown person as unknown, of the methods is calculated. A correct separation occurs if the method does not produce a score above a threshold $T$ in the candidate list of a face query~\cite{grother2017face}. The performance of the methods is calculated separately in the scenarios of clean faces, masked faces, and for faces with the glasses. Then the average performance for all settings is calculated and reported. 

For performance evaluation in closed-set identification, the Cumulative Matching Characteristic (CMC) metric is used. CMC reports the ratio of searches returning the true match at rank $r$ or better. CMC is a non-threshold rank-based metric but in surveillance applications most of the faces are not of any person that is enrolled in the gallery. Therefore, the face of a person that is not enrolled in the gallery should be identified as unknown, which corresponds to the separation performance of the method as mentioned before. The rank-based CMC metric can be extended to a metric with a threshold which ignores the similarity scores that is below the given threshold:

\begin{table}[]
\centering
\caption{Average separation scores of the methods on the custom surveillance dataset.}
\label{tab:sep_scores}
\begin{tabular}{c|c}
\textbf{Method} & \textbf{Separation Performance} \\ \hline
AdaCos & 0.38 \\
AdaM-Softmax & 0.87 \\
AM-Softmax & 0.85 \\
ArcFace & 0.70 \\
CircleLoss & \textbf{0.88} \\
CurricularFace & 0.52 \\
MagFace & 0.84 \\
MV-Softmax & 0.86 \\
NPCFace & 0.77
\end{tabular}
\end{table}

\begin{equation}
F_{match, k} =\dfrac{\sum\limits_{f=0}^{F} \sum\limits_{t=0}^{T_f} B_{k}(f,t)} {\sum\limits_{f=0}^{F} G_f} \qquad k=1,2,..,10
\end{equation}

\[
    B_{k}(f,t)= 
\begin{cases}
    1, & \exists \text{t } \in \text{gallery} \\
    0,              & \text{otherwise}
\end{cases}
\]

\noindent where $F$ corresponds the total frames in the query videos, $k$ is the matching rank, $B_{k}$ is the matching performance calculation function for each person query up to the matching level $k$, $G_f$ is the number of ground-truth bounding boxes of the people in the gallery in the $f$th frame, and $T_f$ is the total number of detections (queries) paired with ground-truth bounding boxes in $f$th frame. We calculated the rank-1 to rank-10 performances of the methods. It is crucial to mention that the RetinaFace detector successfully detects all of the face bounding boxes in the query videos. 

The separation performance of the methods when faces in the query videos are not enrolled in the gallery is calculated as follows:

\begin{equation}
F_{separation} =\dfrac{\sum\limits_{f=0}^{F} \sum\limits_{t=0}^{T_f} A(f,t,e)} {\sum\limits_{f=0}^{F} G_f}
\end{equation}

\[
    A(f,t,e)= 
\begin{cases}
    1,& \text{if } max(t_{\text{scores}}) < \text{e } \text{and }  \forall \text{t } \notin \text{gallery} \\
    0,              & \text{otherwise}
\end{cases}
\]

\noindent where $A$ is the function for calculating the separation performance for each person query, $t$ is the cosine-similarity threshold value for the query. 
We calculate the cosine-similarity scores between the face queries and faces in the gallery. We set similarity threshold $t$ as 0.25 for every method to fairly compare the methods. 

\begin{figure*}
     \centering
     \begin{subfigure}[b]{0.24\textwidth}
         \centering
         \includegraphics[width=\textwidth]{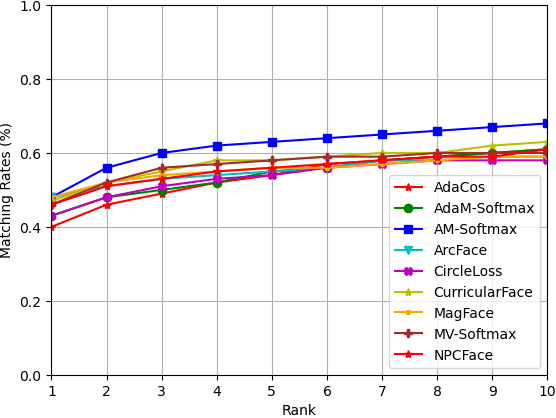}
         \caption{Clean scenario}
         \label{fig:cmc1}
     \end{subfigure}
     \begin{subfigure}[b]{0.24\textwidth}
         \centering
         \includegraphics[width=\textwidth]{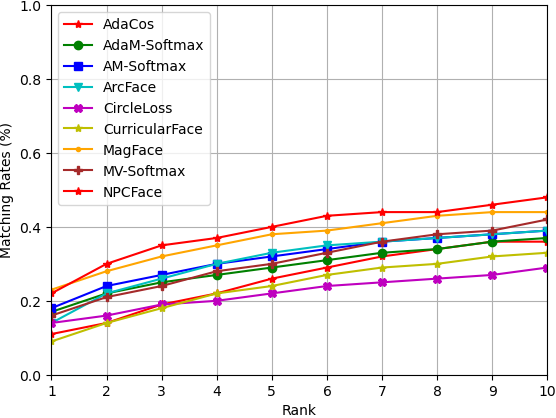}
         \caption{Masked scenario}
         \label{fig:cmc2}
     \end{subfigure}
     \begin{subfigure}[b]{0.24\textwidth}
         \centering
         \includegraphics[width=\textwidth]{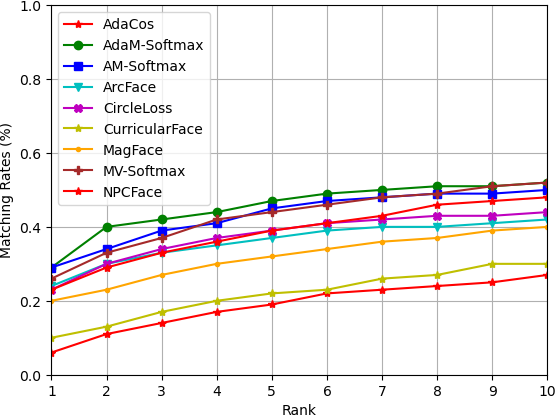}
         \caption{Glasses scenario}
         \label{fig:cmc3}
     \end{subfigure}
      \begin{subfigure}[b]{0.24\textwidth}
         \centering
         \includegraphics[width=\textwidth]{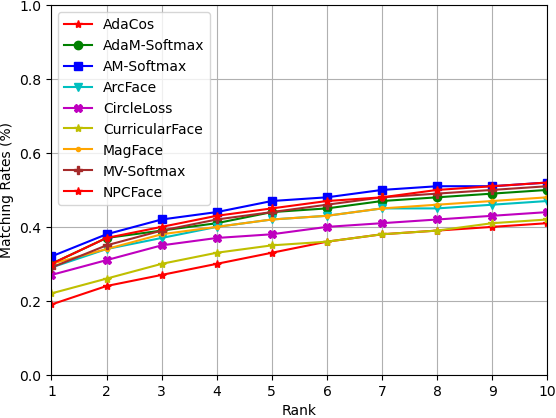}
         \caption{Average scores}
         \label{fig:cmc4}
     \end{subfigure}
        \caption{CMC curves (ranks 1-10) of the methods evaluated on the custom video surveillance dataset.}
        \label{fig:cmc_scores}
\end{figure*}

Table~\ref{tab:sep_scores} shows the separation performance of the methods on our surveillance dataset. In the table, average score is given for the three different scenarios (clean, masked, and glasses). Table~\ref{tab:our_scores} shows the CMC scores of the methods on the video surveillance dataset for the three different scenarios separately and average scores of these scenarios. In clear scenario and total, AM-Softmax achieves the best score for all ranks. For the glasses scenario and the masks scenario, AdaM-Softmax and NPCFace gets the best results, respectively. We can see that different methods achieved the best scores on different face scenarios. One of the reason of this is the models are not trained with facial images that contains glasses or masks. We can also observe from the Figure~\ref{fig:cmc_scores} that AM-Softmax achieves the best performance on the clean faces with a clear margin. Performance gap between methods increase as we introduce glasses and masks to the faces. For example, performance of the CurricularFace and AdaCos methods dropped drastically in glasses scenario. In general, AM-Softmax seems to be more robust to the changes of the face appearance and it also achieves a relatively good performance on separation (Table~\ref{tab:sep_scores}). Although CircleLoss achieved the best separation performance, the matching performance of the method is considerably low.



\begin{table*}[h]
\centering
\caption{CMC scores (ranks 1-10) of the methods with different scenarios on the custom surveillance dataset.}
\label{tab:our_scores}
\begin{tabular}{c|cccccccccc}
\hline
\textbf{Method} & \multicolumn{10}{c}{\textbf{CMC Scores | Scenario: Clean}} \\ \hline
AdaCos & 0.4 & 0.46 & 0.49 & 0.52 & 0.54 & 0.57 & 0.58 & 0.59 & 0.6 & 0.61 \\
AdaM-Softmax & 0.43 & 0.48 & 0.5 & 0.52 & 0.55 & 0.56 & 0.58 & 0.59 & 0.6 & 0.61 \\
AM-Softmax & \textbf{0.48} & \textbf{0.56} & \textbf{0.6} & \textbf{0.62} & \textbf{0.63} & \textbf{0.64} & \textbf{0.65} & \textbf{0.66} & \textbf{0.67} & \textbf{0.68} \\
ArcFace & \textbf{0.48} & 0.51 & 0.53 & 0.54 & 0.55 & 0.56 & 0.58 & 0.58 & 0.59 & 0.59 \\
CircleLoss & 0.43 & 0.48 & 0.51 & 0.53 & 0.54 & 0.56 & 0.57 & 0.58 & 0.58 & 0.58 \\
CurricularFace & 0.47 & 0.52 & 0.55 & 0.58 & 0.58 & 0.59 & 0.6 & 0.6 & 0.62 & 0.63 \\
MagFace & 0.48 & 0.52 & 0.54 & 0.55 & 0.56 & 0.56 & 0.57 & 0.58 & 0.59 & 0.59 \\
MV-Softmax & 0.46 & 0.52 & 0.56 & 0.57 & 0.58 & 0.59 & 0.59 & 0.6 & 0.6 & 0.6 \\
NPCFace & 0.46 & 0.51 & 0.53 & 0.55 & 0.56 & 0.57 & 0.58 & 0.59 & 0.59 & 0.61 \\ \hline
\textbf{Method} & \multicolumn{10}{c}{\textbf{CMC Scores | Scenario: Glasses}} \\ \hline
AdaCos & 0.06 & 0.11 & 0.14 & 0.17 & 0.19 & 0.22 & 0.23 & 0.24 & 0.25 & 0.27 \\
AdaM-Softmax & \textbf{0.29} & \textbf{0.4} & \textbf{0.42} & \textbf{0.44} & \textbf{0.47} & \textbf{0.49} & \textbf{0.5} & \textbf{0.51} & \textbf{0.51} & \textbf{0.52} \\
AM-Softmax & \textbf{0.29} & 0.34 & 0.39 & 0.41 & 0.45 & 0.47 & 0.48 & 0.49 & 0.49 & 0.5 \\
ArcFace & 0.24 & 0.3 & 0.33 & 0.35 & 0.37 & 0.39 & 0.4 & 0.4 & 0.41 & 0.42 \\
CircleLoss & 0.23 & 0.3 & 0.34 & 0.37 & 0.39 & 0.41 & 0.42 & 0.43 & 0.43 & 0.44 \\
CurricularFace & 0.1 & 0.13 & 0.17 & 0.2 & 0.22 & 0.23 & 0.26 & 0.27 & 0.3 & 0.3 \\
MagFace & 0.2 & 0.23 & 0.27 & 0.3 & 0.32 & 0.34 & 0.36 & 0.37 & 0.39 & 0.4 \\
MV-Softmax & 0.26 & 0.33 & 0.37 & 0.42 & 0.44 & 0.46 & 0.48 & 0.49 & 0.51 & 0.52 \\
NPCFace & 0.23 & 0.29 & 0.33 & 0.36 & 0.39 & 0.41 & 0.43 & 0.46 & 0.47 & 0.48 \\ \hline
\textbf{Method} & \multicolumn{10}{c}{\textbf{CMC Scores | Scenario: Masks}} \\ \hline
AdaCos & 0.11 & 0.14 & 0.19 & 0.22 & 0.26 & 0.29 & 0.32 & 0.34 & 0.36 & 0.36 \\
AdaM-Softmax & 0.17 & 0.22 & 0.25 & 0.27 & 0.29 & 0.31 & 0.33 & 0.34 & 0.36 & 0.37 \\
AM-Softmax & 0.18 & 0.24 & 0.27 & 0.3 & 0.32 & 0.34 & 0.36 & 0.37 & 0.38 & 0.39 \\
ArcFace & 0.14 & 0.22 & 0.26 & 0.3 & 0.33 & 0.35 & 0.36 & 0.37 & 0.38 & 0.39 \\
CircleLoss & 0.14 & 0.16 & 0.19 & 0.2 & 0.22 & 0.24 & 0.25 & 0.26 & 0.27 & 0.29 \\
CurricularFace & 0.09 & 0.14 & 0.18 & 0.22 & 0.24 & 0.27 & 0.29 & 0.3 & 0.32 & 0.33 \\
MagFace & \textbf{0.23} & 0.28 & 0.32 & 0.35 & 0.38 & 0.39 & 0.41 & 0.43 & 0.44 & 0.44 \\
MV-Softmax & 0.16 & 0.21 & 0.24 & 0.28 & 0.3 & 0.33 & 0.36 & 0.38 & 0.39 & 0.42 \\
NPCFace & 0.22 & \textbf{0.3} & \textbf{0.35} & \textbf{0.37} & \textbf{0.4} & \textbf{0.43} & \textbf{0.44} & \textbf{0.44} & \textbf{0.46} & \textbf{0.48} \\ \hline
\textbf{Method} & \multicolumn{10}{c}{\textbf{CMC Scores | Total Scores}} \\ \hline
AdaCos & 0.19 & 0.24 & 0.27 & 0.3 & 0.33 & 0.36 & 0.38 & 0.39 & 0.4 & 0.41 \\
AdaM-Softmax & 0.3 & 0.37 & 0.39 & 0.41 & 0.44 & 0.45 & 0.47 & 0.48 & 0.49 & 0.5 \\
AM-Softmax & \textbf{0.32} & \textbf{0.38} & \textbf{0.42} & \textbf{0.44} & \textbf{0.47} & \textbf{0.48} & \textbf{0.5} & \textbf{0.51} & \textbf{0.51} & \textbf{0.52} \\
ArcFace & 0.29 & 0.34 & 0.37 & 0.4 & 0.42 & 0.43 & 0.45 & 0.45 & 0.46 & 0.47 \\
CircleLoss & 0.27 & 0.31 & 0.35 & 0.37 & 0.38 & 0.4 & 0.41 & 0.42 & 0.43 & 0.44 \\
CurricularFace & 0.22 & 0.26 & 0.3 & 0.33 & 0.35 & 0.36 & 0.38 & 0.39 & 0.41 & 0.42 \\
MagFace & 0.3 & 0.34 & 0.38 & 0.4 & 0.42 & 0.43 & 0.45 & 0.46 & 0.47 & 0.48 \\
MV-Softmax & 0.29 & 0.35 & 0.39 & 0.42 & 0.44 & 0.46 & 0.48 & 0.49 & 0.5 & 0.51 \\
NPCFace & 0.3 & 0.37 & 0.4 & 0.43 & 0.45 & 0.47 & 0.48 & 0.5 & 0.51 & 0.52
\end{tabular}
\end{table*}

\section{Conclusion}
In this study, face recognition performance of different face recognition algorithms in surveillance videos is compared under the same settings. For this purpose, we constructed a dataset of face IDs that has high age variance, intra-class variance (face make-up, beard, etc.) with native surveillance facial imagery data for evaluation. Morever, we constructed two synthetic versions of this dataset, masked and glasses versions, to analyze the performance of the methods with occluded faces. We benchmark nine different successful deep learning based face recognition methods for the face identification task on these three scenarios. The experimental results shows that there is a severe decrease in performance when we add glasses or mask to faces. However, we know that the models used are not trained with faces that have glasses or masks. It could be possible to observe a better performance, if we retrain the models with dataset that is augmented with glasses and faces with masks. It is also observable that the performance of the models are very low compared to the scores obtained in facial image benchmarks (such as LFW and MegaFace). So, we can state that it is much more harder to perform identification on surveillance videos and this task is far from being solved.

\textbf{Possible Improvements Specific to Video Surveillance.} To overcome challenges of face recognition in video surveillance, many strategies can be applied. One method would be adding a tracking method to track faces across video frames. After that, these tracklets can be used to identify the individuals by selecting best face sample from the tracklet. The other approach may be discarding the samples that are not suitable for face recognition using the face quality metrics and selecting highest similarity score for identifying the tracked person. Both approaches need some evaluation methods that will help to select the best "quality" face image. Face re-identification methods could also be useful to determine the face ID of a person that has been previously identified with high probability and high fidelity from previous frames. Some of these methods are mentioned in~\cite{cheng2018surveillance,cheng2020face}.




{\small
\bibliographystyle{ieee_fullname}
\bibliography{egbib}
}

\end{document}